\documentclass[sigconf]{acmart}

\AtBeginDocument{%
  }

\usepackage{multirow}
\usepackage{graphicx} 
\usepackage{booktabs}   % 提供三线表支持
\usepackage{makecell}   % 提供单元格内换行支持
\usepackage{subcaption}
\usepackage{pifont}
\usepackage{extarrows}
\usepackage[table]{xcolor} % 颜色支持
\newcommand{\cmark}{\textcolor{green!60!black}{\ding{51}}} 
\newcommand{\xmark}{\textcolor{red!80!black}{\ding{55}}}

\copyrightyear{2026}
\acmYear{2026}
\setcopyright{cc}
\setcctype{by} % TDDO:确认这个(得看erights)
\acmConference[SIGIR '26]
  {Proceedings of the 49th International ACM SIGIR Conference on Research and Development in Information Retrieval}
  {July 20--24, 2026}
  {Melbourne, VIC, Australia.}
\acmBooktitle{Proceedings of the 49th International ACM SIGIR Conference on Research and Development in Information Retrieval (SIGIR '26), July 20--24, 2026, Melbourne, VIC, Australia}
\acmISBN{979-8-4007-2599-9/2026/07}
\acmDOI{10.1145/3805712.3808607}

\begin{document}

\title{From Single- to Cross-Document: Benchmarking Multi-Granularity Event Analysis of Large Language Models}


%%
%作者list
%%
\author{Tao Wen}
\affiliation{
\department{Laboratory of Intelligent Collaborative Computing}
  \institution{University of Electronic Science and Technology of China}
  \city{Chengdu}
  %\state{Sichuan}
  \country{China}
}
\email{enril.wentao@std.uestc.edu.cn}

\author{Shuai Shao}
\affiliation{
\department{Laboratory of Intelligent Collaborative Computing}
  \institution{University of Electronic Science and Technology of China}
  \city{Chengdu}
  %\state{Sichuan}
  \country{China}
}
\email{202221080331@std.uestc.edu.cn}

\author{Pei Ke}
\authornote{Corresponding author.}
\affiliation{
    \department{Laboratory of Intelligent Collaborative Computing}
     \institution{University of Electronic Science and Technology of China}
  \city{Chengdu}
  %\state{Sichuan}
  \country{China}
}
\email{kepei@uestc.edu.cn}

\author{Xu Han}
\affiliation{
    \department{Department of Computer Science and Technology}
  \institution{Tsinghua University}
  \city{Beijing}
  %\state{Beijing}
  \country{China}
  }
\email{han-xu@mail.tsinghua.edu.cn}

\author{Jie Zou}
\affiliation{%
    \department{School of Computer Science and Engineering 
    %(School of Cyber Security)
    }
  \institution{University of Electronic Science and Technology of China}
  \city{Chengdu}
  %\state{Sichuan}
  \country{China}}
\email{jie.zou@uestc.edu.cn}

%\author{Guannan Li}
%\author{Tao Tian}
%\affiliation{
%\department{Laboratory of Intelligent Collaborative Computing}
%  \institution{University of Electronic Science and Technology of China}
%  \city{Chengdu}
%  \state{Sichuan}
%  \country{China}
%}
%\email{lgn4sci@std.uestc.edu.cn}
%\email{taotian@std.uestc.edu.cn}

\author{Guannan Li}
\affiliation{%
\department{Laboratory of Intelligent Collaborative Computing}
  \institution{University of Electronic Science and Technology of China}
  \city{Chengdu}
  %\state{Sichuan}
  \country{China}
  }
\email{lgn4sci@std.uestc.edu.cn}

\author{Tao Tian}
\author{Jinjie Qiu}
\affiliation{%
\department{Laboratory of Intelligent Collaborative Computing}
  \institution{University of Electronic Science and Technology of China}
  \city{Chengdu}
  %\state{Sichuan}
  \country{China}
  }
\email{taotian@std.uestc.edu.cn}
\email{202522900127@std.uestc.edu.cn}

\author{Lan Wang}
\affiliation{
\department{School of Information and Software Engineering}
  \institution{University of Electronic Science and Technology of China}
  \city{Chengdu}
  %\state{Sichuan}
  \country{China}
}
\email{202521090303@std.uestc.edu.cn}
%\author{Jinjie Qiu}
%\affiliation{%
%  \institution{University of Electronic Science and Technology of China}
%  \city{Chengdu}
%  \state{Sichuan}
%\email{qsflower@163.com}

%\author{Lan Wang}
%\affiliation{%
%  \institution{University of Electronic Science and Technology of China}
%  \city{Chengdu}
%  \state{Sichuan}
%\email{202521090303@std.uestc.edu.cn}

\author{Ke Qin}
\affiliation{%
    \department{Laboratory of Intelligent Collaborative Computing}
  \institution{University of Electronic Science and Technology of China}
  \city{Chengdu}
  %\state{Sichuan}
  \country{China}
  }
\email{qinke@uestc.edu.cn}

%这是老版本的作者list
%\author{Tao Wen}
%\author{Shuai Shao}
%\affiliation{%
%  \institution{University of Electronic Science and Technology of China}
%  \city{Chengdu}
%  \state{Sichuan}
%  \country{China}
%}
%\email{202421080222@std.uestc.edu.cn}

%\author{Pei Ke}
%\affiliation{%
%  \institution{University of Electronic Science and Technology of China}
%  \city{Chengdu}
%  \state{Sichuan}
%\email{kepei@uestc.edu.cn}

%\author{Xu Han}
%\affiliation{%
%  \institution{Tsinghua University}
%  \city{Beijing}
  %\state{Beijing}
%\email{thu.hanxu13@gmail.com}

%\author{Jie Zou}
%\affiliation{%
%  \institution{University of Electronic Science and Technology of China}
%  \city{Chengdu}
%  \state{Sichuan}
%\email{jie.zou@uestc.edu.cn}

%\author{Guannan Li}
%\author{Tao Tian}
%\author{Jinjie Qiu}
%\author{Lan Wang}
%\affiliation{%
%  \institution{University of Electronic Science and Technology of China}
%  \city{Chengdu}
%  \state{Sichuan}
%  \country{China}
%}

%\author{Ke Qin}
%\affiliation{%
%  \institution{University of Electronic Science and Technology of China}
%  \city{Chengdu}
%  \state{Sichuan}
%\email{qinke@uestc.edu.cn}

%作者list end****************************

%%\renewcommand{\shortauthors}{Wen et al.}
%要求改成全名了..
\renewcommand{\shortauthors}{Tao Wen et al.}
\begin{abstract}
Event analysis is an essential and fundamental direction of information extraction, involving various event-centric tasks at different granularity of documents. While large language models (LLMs) have preliminarily achieved promising performance in part of these tasks individually, their capability in event analysis still lacks comprehensive understanding due to restricted document granularity, task designs, and data source of existing benchmarks. To address these limitations, we introduce MiGUE-Bench, a systematic benchmark for assessing the performance of LLMs in multi-granularity event analysis. To support large-scale evaluation, we first develop an LLM-driven self-correcting annotation framework called MiGUE-Pipeline, enabling scalable acquisition of high-quality source data of events with automatic labels. Then, we design four core tasks in our benchmark, i.e., event detection, relation reasoning, structure induction, and future prediction, to probe model competence at different levels, from atomic event details to complex cross-document narratives. Extensive experiments on state-of-the-art LLMs and retrieval-augmented generation (RAG) methods delineate the current capability boundary and identify critical deficiencies, providing insights into the future improvement of LLMs in challenging event analysis tasks. 

\end{abstract}





\maketitle
\section{Introduction}
%引入: Event analysis 的定义,范围,重要性
Event analysis is a foundational pillar of information extraction, inherently requiring 
%a multi-granular perspective to transform raw textual data into structured knowledge
to extract the structure of events from raw textual data at different granularity levels
\cite{chen2021event}.
%This paradigm encompasses a hierarchical spectrum of reasoning: 
This essential direction encompasses a hierarchical task taxonomy from the fine-grained event detection of individual occurrences to the induction of global event structures across different documents \cite{eventtask_minard2015semeval,FEP_li2021gragh,eventtask_yang-etal-2025-eventrag,foley2015eventextract,fan2022causal,san2019eventir,lou2022translation,liao2021ed}. %第二个不知道能不能算,是FEP那边用图去做EP的,想法和我们类似但是并非event领域+做的很粗糙%%TODO: 加引用 
Mastering the capabilities of analyzing events from 
%these interconnected scales spanning 
intra-document event mentions to inter-document event relations
is indispensable for the application of current information extraction systems in real-world scenarios,
%artificial intelligence to 
moving beyond local semantic matching towards global relation reasoning.  
%sophisticated events' inference and strategic forecasting in complex real-world scenarios.

%传统方法
Early works on event analysis have extensively explored task-specific deep learning paradigms, ranging from sequence labeling architectures \cite{trade_chen2015event,trade_nguyen2016joint,trade_sha2018jointly} to graph-based reasoning frameworks \cite{FEP_li2021gragh,trade_nguyen2018graph}. Despite their advanced performance on specific datasets, the generalization ability of these models is rather limited. Thus, recent works resort to large language models (LLMs) and preliminarily show their promising performance in part of event analysis tasks, such as event extraction \cite{li2025eesurvey}, event relation prediction \cite{llme_chen2024erl,hu2025llmforere}, and event reasoning \cite{llme_nakshatri2023temporal,tao2025llmer}. Equipped with strong abilities of context understanding and knowledge utilization, LLMs have shown great potential in dealing with complex event analysis tasks in a zero-shot manner, gradually becoming a research focus in this field.

However, we argue that there still lacks a comprehensive benchmark for assessing LLMs' capabilities in event analysis. Existing benchmarks mostly suffer from restricted granularity, lacking data source, and homogeneous task design, hindering a systematic understanding of LLMs' deficiencies at different event-centric tasks:

%%**************重要:加引用
\begin{itemize}
    \item \textit{Restricted Granularity}: Most of the current benchmarks only fall into the assessment at constrained granularity of documents.
    While some of the datasets aim to measure the model performance in understanding events within a document \cite{tao2025llmer,bench_wang2022maven}, others focus on the event relations across multiple documents \cite{hong2016crossdoc,bug2021crossdoc,bench_zhang2024tcelongbench}, both of which fail to provide an entire perspective of LLMs' capabilities at different granularity levels.
    \item \textit{Lacking Data Source}: 
    Most existing benchmarks are retrofitted from legacy datasets \cite{bench_gong2025eventrelbench}, which restrict the scalability to diverse and contemporary corpora. The lack of data sources also causes missing cross-document dependencies, making it improper to measure LLMs' capabilities to analyze events across multiple documents.
    \item \textit{Homogeneous Task Design}: Existing benchmarks are mostly targeting at isolated evaluation dimensions and objects, analyzing the ability of LLMs to deal with specific event relations \cite{bench_zhang2024tcelongbench} or tasks \cite{llme_chen2024erl,llme_nakshatri2023temporal,hu2025llmforere,tao2025llmer}. Such homogeneous and narrow task design may be unable to reveal the capability boundary of general LLMs, exaggerating their performance in the limited task scope.
\end{itemize}

%引入我们的工作MiGUE!
To address these limitations, we introduce \textbf{MiGUE-Bench}, a comprehensive benchmark designed for \textbf{M}ult\textbf{i}-\textbf{G}ran\textbf{U}lar \textbf{E}vent analysis. MiGUE-Bench aims to cover the evolving process of event analysis
%establishes a hierarchical framework, 
from single-document event detection and relation reasoning to cross-document event structure induction and future prediction (as shown in Table \ref{tab:benchmarks}), 
enabling a holistic assessment of LLMs' capabilities in different granularity levels of documents.
%in both single- and cross-document contexts.
\textbf{Firstly}, to deal with the limitation of document granularity and data source, we develop an LLM-driven automatic pipeline named MiGUE-Pipeline that includes document filtering, event / relation annotation, and document clustering, enabling
%that enables 
scalable data construction from raw corpora to support various event-centric tasks. 
%This pipeline operates in several strategic stages, i.e., document filtering, event / relation annotation, and document clustering, ensuring high-quality event-centric data synthesis from raw corpora.
\textbf{Secondly}, to solve the challenge of homogeneous task design, we devise four task types to cover different stages of event analysis, 
%comprehensively probing LLMs' capabilities in distinct evaluation aspects. 
Specifically, \textit{MiGUE-Detection} assesses the performance in 
%perceptual precision in 
recognizing and extracting triggers from event mentions, while \textit{MiGUE-Reasoning} aims to evaluate the capability to reason within and across fragmented document contexts to extract implicit event relations. To further obtain a global understanding of event dynamics across different documents, \textit{MiGUE-Induction} is targeted at evaluating the ability to acquire the global topological structure of events. Finally, \textit{MiGUE-Prediction} tests the capability 
%of predictive modeling, 
%and anticipatory reasoning, 
%challenging LLMs 
to infer the future development of events
%event evolution 
based on the understanding of event dynamics
%local event semantics within a document and global event structures 
across different documents. Our benchmark is expected to 
%provide a comprehensive understanding of LLMs' capabilities 
%in event analysis 
%at different granularity levels and 
reveal the deficiencies of LLMs 
%of LLMs 
via comprehensive and challenging task design.

In summary, our key contributions are three-fold\footnote{The code of MiGUE-Pipeline is released at \url{https://github.com/EnrilAmirite/MiGUE-Pipeline}, while the MiGUE-Bench dataset is available at \url{https://github.com/EnrilAmirite/MiGUE-Bench}.}:

\begin{itemize}
    \item We develop the first open-source 
    %LLM-driven 
    automatic pipeline named MiGUE-Pipeline for 
    %specifically engineered for 
    multi-granularity event data generation, alleviating the dependence on manual annotation and flexibly
    %and enabling the scalable construction of complex cross-document event data to flexibly 
    supporting various event-centric tasks.
    %By leveraging self-correction and relation propagation, our pipeline overcomes the efficiency and consistency bottlenecks of manual annotation, enabling the scalable construction of complex cross-document event networks.
    \item We introduce MiGUE-Bench, a comprehensive benchmark that provides systematic assessment for LLMs' capabilities in the entire spectrum of event analysis.
    \item We conduct extensive experiments on 
    %MiGUE-Bench involving a diverse array of 
    state-of-the-art LLMs and retrieval-augmented generation frameworks, uncovering critical deficiencies of LLMs on event analysis and providing insights for future improvement. 
    %Our findings delineate the current capability boundaries of event-centric modeling, uncovering critical failure modes and pointing toward concrete future research directions.
\end{itemize}
%The code and benchmark are publicly available at \url{https://github.com/EnrilAmirite/MiGUE-Pipeline} and \url{https://github.com/EnrilAmirite/MiGUE-Bench}.

%%**************Bench对比表*****************
%%*******************重要:想最后一个的名字
\begin{table}[!t]
  \caption{Comparison of different event analysis benchmarks.}
  \label{tab:benchmarks}
  \footnotesize
  \centering
  \begin{tabular}{lcccc}
    \toprule
    \textbf{Benchmark} & \begin{tabular}[c]{@{}c@{}}Single-Doc\\ Analysis\end{tabular} & \begin{tabular}[c]{@{}c@{}}Cross-Doc\\ Analysis\end{tabular} & \begin{tabular}[c]{@{}c@{}}Multiple\\Relations\end{tabular} & \begin{tabular}[c]{@{}c@{}}Event\\Structure\end{tabular} \\
    \midrule
    ACE05-EN \cite{Doddington2004ACE}       & \cmark & \xmark & \xmark & \xmark \\
    Causal-TimeBank \cite{mirza2014causal} & \cmark & \xmark & \cmark & \xmark \\
    MAVEN \cite{wang2020maven}          & \cmark & \xmark & \xmark & \xmark \\
    MAVEN-ERE \cite{bench_wang2022maven}      & \cmark & \xmark & \cmark & \xmark \\
    TCELongBench \cite{bench_zhang2024tcelongbench}    & \xmark & \cmark & \xmark & \xmark \\
    EventRelBench \cite{bench_gong2025eventrelbench}   & \cmark & \xmark & \cmark & \xmark \\
    %\cmidrule
    \textbf{MiGUE-Bench (Ours)}     & \cmark & \cmark & \cmark & \cmark \\
    \bottomrule
  \end{tabular}
\end{table}
%%**************Bench对比表 end*****************

\section{Related Work}
\noindent\textbf{LLMs for Event Analysis.}
Recent advances in large language models (LLMs) have enabled a prompt-based paradigm for event-centric understanding and reasoning in a zero-/few-shot setting. Existing studies show that LLMs can support core event analysis capabilities, spanning event extraction \cite{li2025eesurvey,choud2024eaevent,Huang2024textee,li2022dual} and event relation prediction \cite{llme_chen2024erl,hu2025llmforere,Chan2023potential,tao2025llmer}. Beyond event understanding in a single document, LLMs have been utilized for temporal/causal event reasoning 
%and for integrating dispersed evidence 
across documents, commonly with retrieval modules to support more coherent information extraction \cite{llme_nakshatri2023temporal,tao2025llmer,eventtask_yang-etal-2025-eventrag}. To summarize, existing studies suggest that LLMs hold strong potential for bridging multiple granularity of event analysis, ranging from single-document 
understanding to cross-document
%interpretation to higher-level 
reasoning.

Despite this advance, current work on LLM-based event analysis typically focuses on isolated task settings, 
%with specific granularity, data source, ans task design, 
making it hard to delineate the strengths and weaknesses of LLMs along the whole process of event analysis,
%failure modes 
%along the end-to-end pipeline, especially 
moving from local semantics to global structures.
%and downstream forecasting. 
Therefore, a more systematic benchmark is necessary to comprehensively reflect the LLMs' performance at multiple granularity levels of event analysis.

%evaluate LLMs in event analysis.

%借用了一下我之前写的
% Since large language models (LLMs) have demonstrated remarkable capabilities in semantic understanding and reasoning, making them promising candidates for event analysis tasks. A growing body of work has explored the use of LLMs for event extraction \cite{llme_meng-etal-2024-ceanEE}, detection\cite{llme_yan2024collaborateED,llme_10.1007/978-981-97-5492-2_3_ED2} and relation reasoning \cite{llme_chen2024erl,llme_nakshatri2023temporal,llme_luo-etal-2024-openCAU}

\noindent\textbf{Benchmarks for LLMs in Event Analysis.}
Motivated by the recent progress of LLMs on event-centric tasks, several benchmarks have been proposed to probe LLM capabilities along different facets of event analysis. Existing works typically formulate evaluation as instruction-following tasks with structured outputs, aiming to measure whether LLMs can (i) recover event-centric descriptions from long and temporally dense narratives \cite{tao2025llmer,bench_wang2022maven}, (ii) reason about temporal/causal dependencies among events \cite{hong2016crossdoc,bug2021crossdoc,bench_zhang2024tcelongbench}, and (iii) infer event relations under diverse types and conduct event reasoning with different formats
%higher-level event reasoning patterns under diverse relation types and reasoning formats 
\cite{bench_zhang2024tcelongbench,bench_gong2025eventrelbench}. 

Though existing benchmarks offer a preliminary view of LLMs’ event analysis capabilities, they mostly focus on specific event-centric tasks with restricted granularity, data sources, and task design.
%making it difficult to characterize capabilities 
%from local event semantics to global structures 
%in 
%realistic multi-source settings; moreover, benchmark designs are frequently homogeneous, focusing on isolated task formulations or relation types, and thus 
For comparison, our work aims to provide an complete picture of LLM performance involving different stages in the whole process of event analysis, with multiple granularity levels of documents, diverse data sources, and broad task scopes.

%across the full event analysis pipeline.

%借用了一下我之前写的
% With wondering of the effectiveness of LLMs in event analysis, several preliminary evaluation studies have begun to investigate their performance across different event-related tasks. One line of work adopts a document-centric perspective, leveraging document-level event summarization to assess the models' temporal reasoning and narrative comprehension within a holistic context.\cite{bench_zhang2024tcelongbench}.Another line builds on traditional event-centric datasets \cite{bench_wang2022maven,bench_caselli2017eventstoryline,bench_lin2023hieve}and focuses on fine-grained event-relation extraction\cite{bench_gong2025eventrelbench}.

\section{MiGUE-Pipeline}
\begin{figure*}[!t]
  \centering
\includegraphics[width=1.0\linewidth]{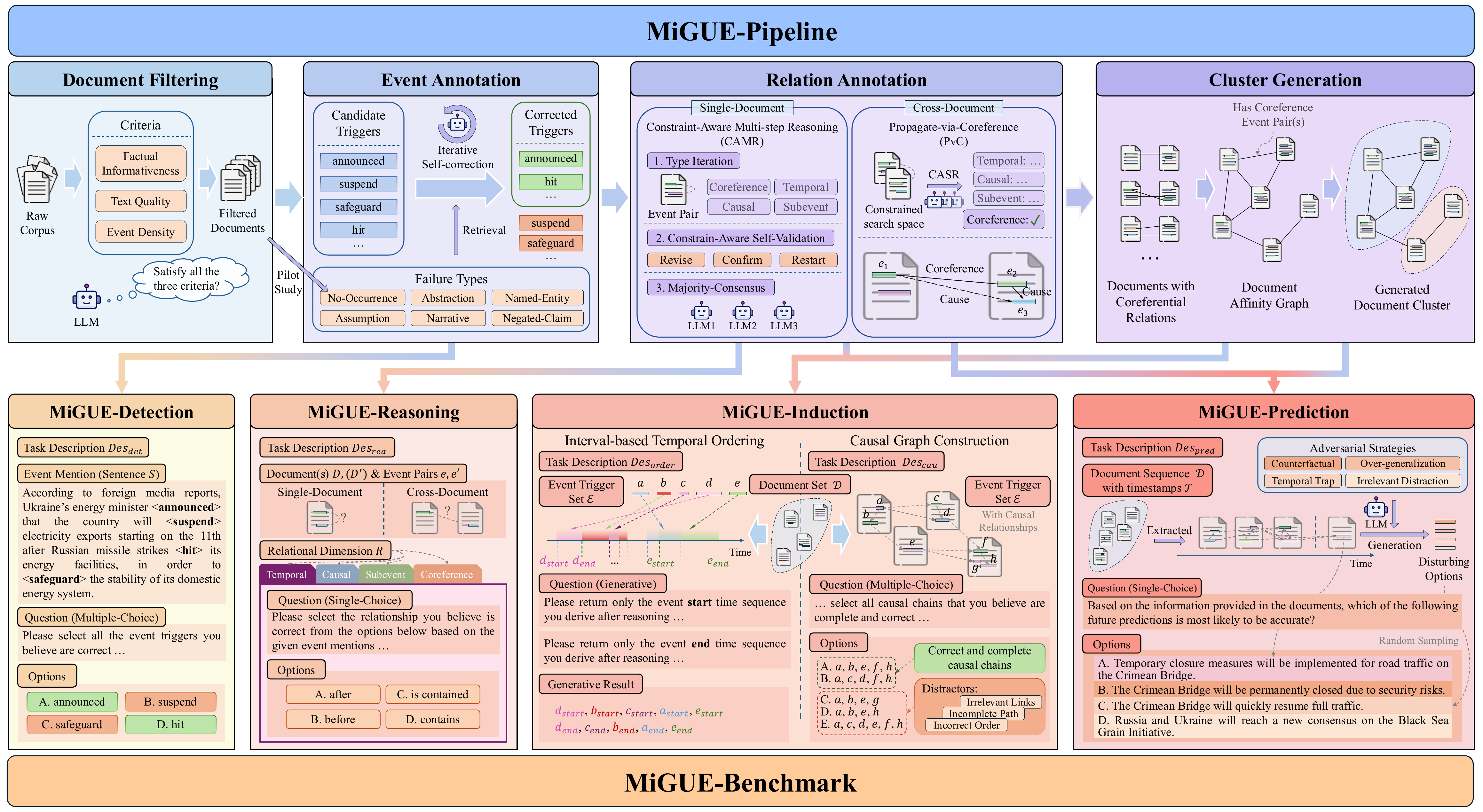}
  \caption{Overview of MiGUE-Pipeline and MiGUE-Bench.}
  %\Description{...}
  \label{fig:overview}
\end{figure*}

To support scalable data construction of MiGUE-Bench, we present 
%an architecture named 
MiGUE-Pipeline, 
%which is engineered as 
a four-stage framework comprising {\itshape document filtering}, {\itshape event annotation}, {\itshape relation annotation}, and {\itshape cluster generation}, as shown in Figure \ref{fig:overview}.
Transforming raw corpora into high-quality multi-granularity event data resources can serve as the foundation for subsequent evaluation.

\subsection{Document Filtering}
\label{sec:docfilter}

%The quality of source documents fundamentally dictates the fidelity of downstream event annotation and relation annotation. 
%To ensure a high-quality data foundation, 
We firstly employ an LLM-driven filtering mechanism that preliminarily evaluates each candidate document across three critical dimensions:
{\itshape factual informativeness} (prioritizing objective reporting over subjective commentary or fragmentary discourse),
{\itshape text quality} (ensuring the documents' grammatical fluency and structural coherence),
{\itshape event density} (excluding redundant or non-informative contents).
A document will be retained only if it satisfies all the aforementioned criteria based on LLM-as-a-Judge with GPT-4o \cite{liu2023geval}. The multi-dimensional filtering ensures that the resulting documents are both grammatically fluent and rich in event dynamics.

\subsection{Event Annotation}
\label{sec:eventannotation}
%***【6种错误类型的定义到时候补在附录】
%好的看来补不了了
Since identifying events in open documents is challenging, we conduct a pilot study on part of documents with GPT-5.2 \cite{GPT-5_systemcard} followed by manual check, finding six typical error types, i.e., {\itshape No-Occurrence, Negated-Claim, Assumption, Abstraction, Named-Entity,} and {\itshape Narrative}. Detailed descriptions of all these error types are provided in Table \ref{tab:event_anno_error}.
Based on these findings, we propose a two-stage protocol for event annotation. Firstly, we make an effective LLM (i.e., DeepSeek-V3.2 \cite{liu2025deepseekv32}) generate a candidate set of event triggers. Then, we devise a retrieval-augmented reflection method to select semantically similar examples in the data pool of each error type from our pilot study, respectively. This method requires the LLM to 
%Third, the model undergoes iterative reflection rounds for each of the six error types. In each round, the LLM is provided with the error definition and the top-3 semantically similar examples retrieved from the case pool. This retrieval-augmented reflection allows the model to 
self-correct the results by reasoning against similar historical failures.
%ultimately leading to better event annotation. 
%补充人工抽样
Manual validation on a subset of 200 data samples shows that our automatic protocol yields a precision of 0.87 during event annotation.
%highlighting the effectiveness of this approach.

%%******************错误类型表格**********
\begin{table}[t]
  \caption{Description of Event Annotation Error Types.}
  \label{tab:event_anno_error}
  \centering
  \footnotesize
  \resizebox{\columnwidth}{!}{
  \begin{tabular}{cp{6.9cm}}
    \toprule
    \textbf{Error Type} & \textbf{Description} \\
    \midrule
    No-Occurrence & {\itshape No-Occurrence} errors arise when the model labels the expressions describing future plans, intentions, or predictions rather than events that have actually occurred as triggers. Although such expressions are event-like in form, they are not realized on the current timeline and should not be extracted as events.
    \\
    \midrule
   Negated-Claim & {\itshape Negated-Claim} errors arise when the model fails to distinguish between the statements about events that did not actually occur and genuinely occurring 
   %speech or action 
   events that express negation (e.g., denying). Negation operators themselves (e.g., not, no, never) do not introduce new events, whereas verbs such as deny, reject, and refute denote individual events and should be identified as triggers. \\
    \midrule
    Assumption & {\itshape Assumption} errors arise when the model identifies as triggers those verbs or action expressions that carry event semantics but occur only in conditional, hypothetical, intentional, advisory, or exhortative contexts, rather than being asserted as factual events on the real timeline. 
    %Such expressions should not be annotated as valid trigger words in event detection.
    \\
    \midrule
    Abstraction & {\itshape Abstraction} errors arise when the model labels the expressions that are event-like in form but function rhetorically or metaphorically rather than denote a specific real-world event as triggers. Such expressions lack clear temporal anchoring, identifiable participants, and concrete boundaries. Thus, they are not extractable factual events.\\ 
    \midrule
    Named-Entity & {\itshape Named-Entity} errors arise when the model incorrectly labels nominal elements of an event (e.g., participants, carriers, results, named entities, event-denoting nouns, or state descriptions) as triggers. %However, valid triggers should be predicate-like expressions that explicitly signal the occurrence of an event. 
    %These nominal elements function only as event arguments rather than encoding the event occurrence itself, and thus should not be annotated as trigger words. 
    \\
    \midrule
    Narrative & {\itshape Narrative} errors arise when the model labels the expressions that narrate, explain, modify, or evaluate an event rather than denote the core action as event triggers. Such expressions (e.g., stance and manner) do not introduce new events but only supplement or interpret existing ones. Thereby, they should not be annotated as events.
    \\
    \bottomrule
  \end{tabular}}
\end{table}
%%******************错误类型表格 end**********

\subsection{Relation Annotation}
\label{sec:relannotation}

To further acquire the relation between the events within a single document or among different ones, we devise different strategies to ensure the quality of automatic relation annotation.

\noindent\textbf{Single-Document Relation Annotation.}
%Building upon the \cite{llme_chen2024erl} framework, 
Inspired by existing works \cite{llme_chen2024erl}, 
%we implement a Constraint-Aware Step-wise Reasoning strategy to capture complex event dependencies. 
we design a constraint-aware multi-step reasoning strategy to capture event dependencies within a document. 
%This approach treats relation annotation as an iterative decision process characterized by dynamic constraint pruning and multi-model consensus.
For each event pair, LLMs iteratively select the most confident relation type. Upon selecting a candidate label at each iteration, LLMs are prompted to validate its decision against a set of explicit logical constraints (e.g., a coreference label inherently precludes causal or sub-event relations). Based on self-evaluation, the LLM should select one of the following operations, i.e., 
%must: 
%{\itshape confirm} (
\textit{confirming} the current label 
%as logically sound 
to proceed to the next most confident relation type, 
%{\itshape revise} (identifying a discrepancy in 
\textit{revising} the current label to select an alternative one, and
%), and 
{\itshape restarting} the entire decision process due to the conflict during reasoning.
%(detecting a conflict during reasoning and deciding a full reset of the entire decision process). 
Once a label is confirmed, we dynamically prune the search space by removing logically incompatible relation types. 
%ensuring the global logical consistency of the annotation.
%
To further enhance reliability, we employ majority voting with three LLMs (i.e., GPT-4o, DeepSeek-V3.2 \cite{liu2025deepseekv32}, and Qwen3-Max \cite{yang2025qwen3}). If all the models yield identical predictions, the result is accepted; otherwise, the instance is escalated to a high-capacity LLM (e.g., Claude-4.5-Opus) for final results.
%补充抽样数据
This approach achieves
%In a sampled evaluation, this approach achieved 
an accuracy rate of 0.82 on 100 data samples during manual check, which shows its effectiveness.

%%************主图**********************
%%************主图**********************

\noindent\textbf{Cross-Document Relation Annotation.}
%Different from single-document settings, 
For comparison, cross-document relation annotation faces the challenges of relational sparsity and computational intractability.
%is a more difficult task, which 
%faces two primary challenges, i.e., \textit{relational sparsity} (where cross-document event pairs mostly lack explicit dependencies) and \textit{computational intractability} (where exhaustive pairing across a large amount of documents leads to a prohibitively vast search space).
To address these challenges, we propose a propagate-via-coreference strategy. For preparation, 
%Firstly, 
we constrain the search space using three proximity heuristics, including {\itshape temporal proximity} (where documents must fall within a one-month timestamp window), {\itshape semantic affinity} (where event mentions should exceed a semantic similarity threshold), and
%(set at 0.7). 
{\itshape entity overlap} (where events should share common participants).
Potentially coreferential pairs are then validated using the same protocol as in the single-document part.
In the propagate-via-coreference strategy, we densify the relation network by treating coreferential events as bridges for propagation. By applying formal transitivity rules (e.g., if $e_{1}\xlongequal{coref}e_{2}$, $e_{1} \xrightarrow{cause} e_{3}$, then $e_{2}\xrightarrow{cause} e_{3}$),
we recover implicit dependencies across different documents. 
This systematic approach circumvents exhaustive pairing while improving the global structural integrity of the event relation network.

\subsection{Cluster Generation}
\label{sec:clustergen}

%Recognizing that 
Since documents that share identical real-world events are potentially
%are fundamentally 
linked by their underlying narrative contents,
%logical and narrative structures, 
we utilize event coreference as a primary signal for document cluster generation, supporting the data construction of downstream tasks that require sufficient cross-document relations.
%corpus organization. 
Specifically, we construct a document affinity graph whose vertices represent individual documents and edges denote the presence of coreferential event pairs. To partition this graph into dense subgraphs,
%manageable, high-potential units, 
we employ the Leiden community detection algorithm \cite{traag2019leiden}, which isolates clusters with maximum narrative coherence. By operating on these localized subgraphs whose relational density is highest, we streamline the discovery of closely-connected documents for more efficient data construction of cross-document event-centric tasks.

\section{MiGUE-Benchmark}
In this section, we systematically delineate the construction of MiGUE-Bench, including four core tasks.
%a multi-granularity benchmark for measuring LLMs on a spectrum of event-centric capabilities. 
%Drawing from the source data generated in the MiGUE-Pipeline, we formulate four distinct tasks that transition from local event detection to global narrative synthesis. 
For each task
% of MiGUE-Bench
, we provide a rigorous definition involving task inputs and outputs, followed by 
its 
% the specific 
data construction strategy with MiGUE-Pipeline. 
%which enables efficient generation 
%used to translate 
%from raw relational data into 
%of high-quality evaluation data instances.

\subsection{MiGUE-Detection}

Since the ability of precisely detecting events is the foundational skill of event analysis, we devise the MiGUE-Detection task to measure LLMs' performance on distinguishing factual event triggers.

\noindent\textbf{Task Definition.}
Given a task description $Des_{det}$, a sentence $S$, and a candidate set of event triggers $\mathcal{C} =\left\{ c_{1},c_{2},...,c_{r} \mid 3\leqslant r\leqslant 6\right\}$ where $r$ indicates the size of $\mathcal{C}$, the model is required to provide all the valid triggers that appear in $S$ within the candidate set $\mathcal{C}$.

\noindent\textbf{Data Construction.}
The data instances for this task are directly acquired from the event annotation stage (Section \ref{sec:eventannotation}). To further improve the difficulty of this fundamental task, we incorporate
%To transform these into a rigorous test of discriminatory power, we move beyond simple identification by incorporating 
adversarial distractors into the options. Specifically, 
% in addition to the correct option, 
%for each sentence, 
we inject typical error types 
%(e.g., Negated Events, Abstractions) 
identified during event annotation of MiGUE-Pipeline into candidate options for each instance. By integrating these failures modes 
as distractors, we aim to assess the fine-grained semantic understanding ability of LLMs.
%compel the model to execute fine-grained semantic and logical discrimination. 

%%******************Benchmark统计表*********
\begin{table}[t]
  \caption{Statistics of MiGUE-Bench including the number of data instances (\#Instance) and the average number of input tokens (\#Token) / options (\#Option).}
  \label{tab:dataset_stats_vertical}
  \centering
  \footnotesize
  \begin{tabular}{cll ccc}
    \toprule
    \textbf{Task} & \multicolumn{2}{c}{\textbf{Subtask}} & \textbf{\#Instance} & \textbf{\#Token} & \textbf{\#Option} \\
    \midrule
    Detection & \multicolumn{2}{c}{-} & 330 &55.18 &4.56  \\
    \midrule
    \multirow{8}{*}{Reasoning} & \multirow{4}{*}{Intra} & Temporal & 300 &796.84 & 4.00 \\
    & & Causal & 300 &937.31 & 4.00\\
    & & Coreference & 300 & 890.00& 2.00\\
    & & Subevent & 300 & 770.58& 3.00\\
    \cmidrule(lr){2-6}
    & \multirow{4}{*}{Cross} & Temporal & 300 &1209.37 & 4.00 \\
    & & Causal & 300 & 1202.18& 4.00 \\
    & & Coreference & 300 &1202.24 & 2.00 \\
    & & Subevent & 300 &1141.26 & 3.00 \\
    \midrule
    \multirow{2}{*}{Induction} & \multicolumn{2}{c}{Temporal Order} & 150 &671.77 & -  \\
    & \multicolumn{2}{c}{Causal Graph} & 110 &1431.27 & 4.48\\
    \midrule
    Prediction & \multicolumn{2}{c}{-} & 300 &2112.03 &4.50 \\
    \midrule
    Overall & \multicolumn{2}{c}{-} & 3,290 & 1061.24&3.85 \\
    \bottomrule
  \end{tabular}
\end{table}
%%******************Benchmark统计表 end*********

\subsection{MiGUE-Reasoning}
%这是第四版
Inferring relationships between events within / across documents is essential for event analysis.
%structure under.
%deciphering the event structures at multiple granularity levels. 
%structural logic of complex narratives. 
MiGUE-Reasoning focuses on evaluating LLMs' capacity of reasoning across different events.
%event relations across both single-document and cross-document contexts.

\noindent\textbf{Task Definition.}
Given the task description $Des_{rea}$, the event pair $(e, e^{'})$ which comes from the document $D / D^{'}$, respectively, the target relational dimension $\mathcal{R}\in \{Temporal,Causal,Subevent,Co-reference\}$, and a candidate set of relation labels $\mathcal{C} =\{ c_{1},c_{2},...,c_{r}\mid 2\leqslant r\leqslant 4\}$, the model is required to select the a label from $\mathcal{C}$ that precisely characterizes the relationship within the corresponding dimension. According to whether $D$ and $D^{'}$ are the same document, this task can be further categorized into two subtasks, i.e., \textit{Reasoning-Intra} and \textit{Reasoning-Cross}.

\noindent\textbf{Data Construction.}
To ensure that our benchmark probes deep reasoning over events rather than surface-level pattern matching, 
%we curate instances from our pipeline's annotations with two rigorous filters:
%{\itshape Implicit Relation Filtering}: 
we explicitly exclude pairs linked by overt linguistic cues, such as direct temporal markers (\textit{afterward}) or causal connectives (\textit{because}). We retain only those relations that necessitate contextual understanding and multi-hop reasoning.
%{\itshape Adversarial Hard Cases}: 
We also prioritize event pairs that the effective LLMs (such as GPT-4o) initially misclassify to further enhance the difficulty.
%By incorporating these cases, we transform the task into a diagnostic probe for the limits of current LLM reasoning.

\subsection{MiGUE-Induction}
Global comprehension of event development necessitates inducing the topological structure of event across documents. MiGUE-Induction aims to measure the capacities of LLMs for structure induction through two sub-tasks:
{\itshape Interval-based Temporal Ordering} and
{\itshape Causal Graph Construction}.

\subsubsection{Interval-based Temporal Ordering}

Traditional temporal ordering often simplifies events into instantaneous points, relying on binary before/after comparisons \cite{bench_zhang2024tcelongbench}. This task design fails to capture the temporal duration and interval logic (e.g., overlap and containment) inherently in real-world narratives. To address this limitation, we decouple each event into its \textit{start} and \textit{end} temporal anchors for fine-grained ordering. 

\noindent\textbf{Task Definition.}
%By requiring models to order these boundaries independently, we move beyond sequencing to assess a model’s temporal span awareness and its ability to reconstruct the relative duration and overlap of multiple events.
%
Given the task description $Des_{order}$, a set of event triggers $\mathcal{E}=\{e_1,\dots,e_n \mid n\geq 3\}$, and a set of documents $\mathcal{D}=\{D_1,\dots,D_m \mid m\geq 2\}$ where these events occur, the model should 
%
%For each instance, the model is provided with: The task description $des_{order}$. A set of triggers to be ordered $\{\mathcal{E}={e_1,\dots,e_n}\mid n\geq 3\}$. And background documents $\{\mathcal{D}={D_1,\dots,D_m}\mid m\geq 2\}$ in which these events occur. In this cross-document setting, the model must 
output two distinct permutations of $\mathcal{E}$, representing the chronological order of event starts and event ends, respectively.

\noindent\textbf{Data Construction.}
The data instances 
%for this task 
are derived from cross-document temporal relations (Section \ref{sec:relannotation}) and document clusters (Section \ref{sec:clustergen}) generated by MiGUE-Pipeline. We specifically leverage the results of relation propagation, which yields dense temporal networks across different documents. 
%To transform these raw annotations into a rigorous test of the ability to distinguish events' temporal orders,
we prioritize clusters with high relation density, ensuring that the selected events form an interconnected network rather than isolated chains.
%{\itshape Timestamp Stripping}: 
Furthermore, to prevent the models from relying on superficial pattern matching, we remove absolute temporal markers (e.g., April 14, 2022) in the data instances.
%from a subset of instances. 
This forces the model to induct the relative temporal order
%—and the resulting onset/offset permutations—
from the narrative contents and implicit cues within the provided documents.

%\subsubsection{DAG-based Causal Induction}
\subsubsection{Causal Graph Construction}

Causality among events of different documents commonly manifests as a directed acyclic graph (DAG), where events may have multiple preconditions or cause several subsequent consequences. Since causality is essential for structure induction of events, we design a task to assess the capability of constructing causal graphs.

\noindent\textbf{Task Definition.}
%Real-world causality is seldom linear; it manifests as a Directed Acyclic Graph (DAG) where events may have multiple preconditions or cause several subsequent consequences. 
Given the task description $Des_{cau}$, a set of event triggers $\mathcal{E}=\{e_1,...,e_n\mid n\geq 3\}$, a set of documents $\mathcal{D}=\{D_1,...,D_m \mid m\geq 2\}$ where these events occur, and a set of candidate options $\mathcal{C}_{s_{i}} =\{ c_{1},c_{2},\dots,c_{r}\mid 4\leqslant r\leqslant 6\}$, where
each option corresponds to a causal path $\langle e_1\rightarrow e_2\rightarrow \dots \rightarrow e_n \rangle$, the model is required to identify all the longest valid causal chains within the causal DAG.
%—defined as the longest possible paths within the causal DAG. 
%This requires the model to synthesize evidence across documents to reconstruct the complete causal topology rather than identifying isolated pairs.

\noindent\textbf{Data Construction.}
%Instances are curated from the causal networks generated by the MiGUE-Pipeline.
The data instances are curated from causal networks that are built based on causal relations (Section \ref{sec:relannotation}) and cluster generation (Section \ref{sec:clustergen}).
To acquire
%transform these Causal DAGs into 
challenging evaluation instances, we treat each maximal path in causal DAGs as a correct option and design three types of distractors as disturbing options, including {\itshape irrelevant links} (i.e., chains containing event pairs with no logical dependency), {\itshape incomplete path} (i.e., sub-paths that are causally valid but fail to include all possible intermediate or terminal events)
%, testing the model's completeness of reasoning
, and {\itshape incorrect order} (i.e., chains where events are causally related but presented in an incorrect logical order).

%最后修一下后面几行...
%%*******************主表*********************
%gemini说要在导言区放两个宏包(放了已经
% \usepackage{multirow}
% \usepackage{graphicx} 

\begin{table*}[t]
  \caption{Accuracy (Acc.) and Micro-F1 of different LLMs and RAG methods on MiGUE-Benchmark.}
  \label{tab:main_results}
  \centering
  \footnotesize
  \setlength{\tabcolsep}{3.8pt} 
\begin{tabular}{l|c|cccc|cccc|cc|c|c}
    \toprule
    % 使用 multirow{4} 并配合 * 让其在 4 行表头中垂直居中
     \textbf{Granularity} & \multicolumn{5}{c|}{\textbf{Single-Document}} & \multicolumn{8}{c}{\textbf{Cross-Document}} \\
    \cmidrule(lr){1-1} \cmidrule(lr){2-6} \cmidrule(lr){7-14}
    \textbf{Task} & \textbf{Detection} & \multicolumn{8}{c|}{\textbf{Reasoning}} & \multicolumn{3}{c|}{\textbf{Induction}} & \textbf{Prediction} \\
    \cmidrule(lr){1-1}\cmidrule(lr){2-2} \cmidrule(lr){3-10} \cmidrule(lr){11-13} \cmidrule(lr){14-14}
    \multirow{2}{*}{\textbf{Subtask}} & \multirow{2}{*}{\textbf{-}} & \multicolumn{4}{c|}{\textbf{Reasoning-Intra}} & \multicolumn{4}{c|}{\textbf{Reasoning-Cross}} & \multicolumn{2}{c|}{\textbf{Temporal Order}} & \multirow{2}{*}{\textbf{Causal Graph}} & \multirow{2}{*}{\textbf{-}} \\
    \cmidrule(lr){3-6} \cmidrule(lr){7-10} \cmidrule(lr){11-12}
    & & \textbf{Temp.} & \textbf{Cau.} & \textbf{Coref.} & \textbf{Sub.} & \textbf{Temp.} & \textbf{Cau.} & \textbf{Coref.} & \textbf{Sub.} & \textbf{Start} & \textbf{End} & & \\
        \cmidrule(lr){1-1}\cmidrule(lr){2-2} \cmidrule(lr){3-10} \cmidrule(lr){11-13} \cmidrule(lr){14-14}
    \textbf{Metric} & \textbf{Micro-F1} & \textbf{Acc.} & \textbf{Acc.} & \textbf{Acc.} & \textbf{Acc.} & \textbf{Acc.} & \textbf{Acc.} & \textbf{Acc.} & \textbf{Acc.} & \textbf{Acc.} & \textbf{Acc.} & \textbf{Acc.} & \textbf{Acc.} \\
    \midrule
    \multicolumn{14}{c}{\textit{Closed-Source LLMs}} \\
    \midrule
    GPT-5.2-Pro & 0.8398 & 0.6600 & 0.6373 & \textbf{0.9735} & 0.8033 & 0.6786 & 0.6408 & \textbf{0.9636} & 0.8188 & 0.8412 & \textbf{0.8333} & 0.4375 & 0.4231 \\
    Gemini-3-Pro & 0.7151 & 0.6867 & 0.6296 & 0.9500 & 0.7801 & 0.7967 & 0.5571 & 0.9100 & 0.8034 & 0.6623 & 0.3775 & 0.2159 & 0.5814 \\
    Claude-4.5-Opus & \textbf{0.8552} & 0.7100 & \textbf{0.6778} & 0.9700 & \textbf{0.8289} & 0.7781 & \textbf{0.7048} & 0.9333 & 0.8421 & \textbf{0.8742} & 0.8079 & \textbf{0.5814} & 0.5544 \\
    Claude-4.5-Haiku & 0.6101 & 0.6767 & 0.6000 & 0.9300 & 0.7000 & 0.4633 & 0.4238 & 0.7400 & 0.7164 & 0.4437 & 0.1722 & 0.1932 & 0.4452 \\
    Qwen3-Max & 0.6543 & 0.6967 & 0.6037 & 0.9050 & 0.8022 & 0.7067 & 0.6000 & 0.7700 & 0.8099 & 0.4437 & 0.1987 & 0.2727 & 0.4585 \\
    \midrule
    \multicolumn{14}{c}{\textit{Open-Source LLMs}} \\
    \midrule
    DeepSeek-V3.2 & 0.8168 & 0.6600 & 0.6630 & 0.9625 & 0.7956 & 0.6732 & 0.6286 & 0.8567 & \textbf{0.8480} & 0.8211 & 0.6300 & 0.2841 & 0.5210 \\
    GLM-4.7 & 0.8133 & 0.6933 &	0.6167 & 0.9633 & 0.7766 & 0.7066 & 0.5333 & 0.8600 & 0.8300 & 0.8067 & 0.5600 & 0.2727 & 0.4867  \\
    Kimi-K2 & 0.5894 & 0.5900 & 0.5630 & 0.9200 & 0.7044 & 0.6167 & 0.5143 & 0.8400 & 0.7807 & 0.3576 & 0.1987 & 0.2614 & 0.5150 \\
    Qwen3-8B & 0.6138 & 0.4000 & 0.4599 & 0.8677 & 0.5133 & 0.4900 & 0.3390 & 0.7333 & 0.6170 & 0.1391 & 0.0795 & 0.1364 & 0.3355 \\
    Qwen3-30B & 0.6298 & 0.4100 & 0.4926 & 0.6211 & 0.6022 & 0.5372 & 0.5095 & 0.6200 & 0.7895 & 0.1722 & 0.0530 & 0.2500 & 0.3522 \\
    Qwen3-235B & 0.6115 & 0.5600 & 0.5481 & 0.8725 & 0.6133 & 0.5567 & 0.5286 & 0.7667 & 0.8362 & 0.3510 & 0.1258 & 0.1705 & 0.4352 \\
    Llama-3.1-7B & 0.4959 & 0.2326 & 0.2185 & 0.3296 & 0.2238 & 0.2167 & 0.1667 & 0.1800 & 0.1316 & 0.0596 & 0.0132 & 0.0341 & 0.3654\\
    \midrule
    \multicolumn{14}{c}{\textit{RAG Methods based on LLMs}} \\
    \midrule
 %\multirow{3}{*}{\makecell[l]{UltraRAG \\ (Vanilla RAG)}}  Gemini-3-pro 
 UltraRAG (w/ Gemini-3-Pro)
 & 0.7457 & 0.6939 & 0.6148 & 0.9525 & 0.8263 & \textbf{0.8000} & 0.5333 & 0.8900 & 0.8304 & 0.6954 & 0.3841 & 0.2045 & \textbf{0.6246} \\
    UltraRAG (w/ Qwen3-Max) &0.6936 & 0.7000 &0.6074 &0.9100 &0.7911 &0.7067 &0.6190 &0.7900 &0.7953 &0.4768 &0.2185 &0.2955 &0.4751\\
    UltraRAG (w/ Qwen3-8B) &0.6287&0.3667 &0.4333 &0.8690 &0.5178 &0.5047 &0.3301 &0.6900 &0.5409 &0.1457 &0.0728 &0.1136 &0.3522\\
    %\cmidrule(lr){1-15} %
    %\multirow{3}{*}{\makecell[l]{LightRAG}} 
    LightRAG (w/ Gemini-3-Pro)	&0.7137 &0.7100 &0.6181 &0.9550 &0.7328 &0.7500 &0.5571 &0.8467 &0.7895 &0.7033 &0.4196 &0.2067 &0.6185\\
    LightRAG (w/ Qwen3-Max) &0.6465 &\textbf{0.7181} &0.6132 &0.8922 &0.7497 &0.6967 &0.6238 &0.7933 &0.7924 &0.4834 &0.2318 &0.2955 &0.4950\\
LightRAG (w/ Qwen3-8B) &0.5851 &0.3893 &0.4481 &0.8647 &0.5145 &0.4867 &0.4012 &0.6467 &0.5380 &0.1126 &0.0397 &0.1277 &0.3445\\
    \bottomrule
  \end{tabular}
\end{table*}

%%*******************主表 end*******************

\subsection{MiGUE-Prediction}
The purpose of event analysis is not merely the retrospective understanding of what has occurred, but also the prediction of what will follow. Thus, we devise the 
%The hallmark of event intelligence lies in prospective anticipation. 
MiGUE-Prediction task as follows.
%that focuses on the assessment of LLMs' forecasting ability.
%through causal and temporal extrapolation.

\noindent\textbf{Task Definition.}
Given the the task description $Des_{pred}$, a temporally ordered document sequence $\mathcal{D}=\{D_1,...,D_{m-1}| 3\leqslant m\leqslant 6\}$ with their timestamps, %$\mathcal{T}=\{t_1,...,t_{m-1}|3\leqslant m\leqslant 6\}$
and a candidate set of events $\mathcal{C} =\{c_{1},\dots,c_{r}\mid 4\leqslant r\leqslant 5\}$, the model is required to select the event that occurs in the correct terminal document $D_{m}$ from $\mathcal{C}$.

%For each question$Q_{EP}$, we present the model with: A task description $des_{EP}$. A chronologically ordered reference document sequence with their explicit publication timestamps$\{\mathcal{D}=\{D_1,...,D_{m-1}\}, \mathcal{T}=\{t_1,...,t_{m-1}\}\mid 3\leqslant m\leqslant 6\}$. And a candidate set $\mathcal{C}_{s_{i}} =\{ c_{1},c_{2},\dots,c_{r}\mid 4\leqslant r\leqslant 5\}$, containing the ground-truth event description from the terminal document $D_{m}$ and several reasoning-based distractors. The model must synthesize the latent causal threads to identify the actual evolution of the narrative.

\noindent\textbf{Data Construction.} 
Inspired by existing works \cite{FEP_huang2024causal,FEP_li2021gragh,FEP_ma2023SeCoGD}, we extract the document sequences from the clusters (Section \ref{sec:clustergen}) 
%in MiGUE-Pipeline 
that exhibit cross-document causal chains (Section \ref{sec:relannotation}). 
%To ensure the data quality, 
%each sequence is filtered by Deepseek-V3. based on information richness, topical coherence, and causal relevance.
%(the dependency of $D_m$ on preceding events). 
We further increase difficulty by appending redundant context—chronologically consistent but irrelevant documents from the same cluster—to discourage shallow pattern matching. For option design, we devise the correct option 
as a paraphrased version of the event occurring in $D_m$
%$e_{sink}$ 
with specific details removed to prevent trivial lexical shortcuts.
%To rigorously test the model's logic and 
Following existing works \cite{FEP_guan2024openep}, we generate disturbing options using four adversarial strategies, including
{\itshape counterfactual} (i.e., events that are logically opposite to the true outcome),
{\itshape overgeneralization} (i.e., plausible but exaggerated conclusions that transcend the evidence),
{\itshape temporal trap} (i.e., past events from the document sequence), and 
%testing the model's ability to distinguish history from future.
{\itshape irrelevant distraction} (i.e., fabricated but contextually consistent events unsupported by the document sequence).

%\subsection{Quality Control and Ethical Safety}
\subsection{Quality Control}
To ensure the quality of MiGUE-Bench, we implement a multi-stage validation protocol. Firstly, we automatically filter all the instances to 
%{\itshape Structural and Source Diversity}: instances are systematically filtered to 
maintain a balanced distribution across document sources, event densities, and difficulties. 
Then, we thoroughly
%{\itshape Content Filtering}: a strict safety pipeline is applied to 
detect and exclude offensive, biased, or harmful content, ensuring
% that this benchmark aligns 
alignment with ethical standards.
%while providing a high-fidelity, safe environment for evaluating complex event reasoning.
Finally, we conduct manual review on all the instances
%{\itshape Expert Verification}: expert review is conducted on a sampled subset 
to verify the correctness of answers
%ground-truth accuracy 
and eliminate ambiguity in all the options. The statistics of the final dataset are shown in Table \ref{tab:dataset_stats_vertical}.

\section{Experiment}

\subsection{Setting}
\noindent\textbf{Source Data.}
%EDIT:source 来源
As MiGUE-Pipeline is a scalable event data construction pipeline, 
%which supports annotation multi-linguistic corpus.
our corpus is mainly drawn from two sources: (1) high-quality open-source news document datasets \cite{corpus_ma2023structured}, and (2) more than 9,000 news reports collected from Chinese official and mainstream media outlets between 2022 and 2024. During data collection, we performed a preliminary filtering based on news tags to ensure that the corpus covers major events from this period as comprehensively as possible. All the corpora are subsequently processed by our MiGUE-Pipeline to automatically construct event data\footnote{Since the source data collected is initially Chinese, we use MiGUE-Pipeline to construct the Chinese dataset.
Nevertheless, MiGUE-Pipeline is language-agnostic and can be also applied to other languages with corresponding prompts.}.
%To improve the generality of MiGUE-Bench, we will provide the English-translated version with careful manual check.}.

%原文
%Our source data consists of existing public datasets \cite{corpus_ma2023structured} together with major news articles (2022–2024) crawled from diverse media platforms. All the corpora are subsequently processed by our MiGUE-Pipeline to automatically construct event data
%\footnote{Since the source data collected is initially Chinese, we use MiGUE-Pipeline to construct the Chinese dataset.
%Nevertheless, MiGUE-Pipeline is language-agnostic and can be also applied to other languages. To improve the generality of MiGUE-Bench, we will provide the English-translated version with careful manual check.}.

\noindent\textbf{Base Model Selection.}
%从开闭源说
To comprehensively evaluate the performance of LLMs, we select two lines of work, i.e., LLMs and RAG methods. For LLMs, we involve mainstream closed-source (including GPT-5.2-Pro \cite{GPT-5_systemcard}, Gemini-3-Pro \cite{google_gemini_3_2025}, Claude-4.5-Opus \cite{anthropic_claude_opus_4.5_2025}, Claude-4.5-Haiku \cite{anthropic_claude_haiku_4.5_2025}, and Qwen3-Max \cite{yang2025qwen3}) and open-source LLMs (including DeepSeek-V3.2 \cite{liu2025deepseekv32}, GLM-4.7 \cite{zeng2024chatglm}, Kimi-K2 \cite{team2025kimi}, Qwen-3-8B/30B/235B \cite{yang2025qwen3}, and Llama-3.1-7B \cite{grattafiori2024llama}) as representatives, covering the models with different scales, families, and capacities. Furthermore, we also 
%measure the performance of LLMs under a RAG setting, where we 
adopt two mainstream RAG frameworks including LightRAG \cite{rag_guo2025lightrag} and UltraRAG \cite{rag_chen2025ultrarag}. The retrieval corpora contain 5,000 documents from the document filtering stage (Section \ref{sec:docfilter}) of MiGUE-Pipeline\footnote{In the task of MiGUE-Prediction, we restrict the documents to those occurring prior to the predicted event, preventing data leakage.}. In our experiment, we utilize Gemini-3-Pro, Qwen3-Max, and Qwen3-8B as base models for RAG methods.

\noindent\textbf{Evaluation Metric.}
%除了Detection其他的都是acc
For MiGUE-Detection, we adopt Micro-F1 to jointly reflect the precision and recall of this multi-choice selection task. As for the other tasks, accuracy is adopted for measurement.

\subsection{Main Result}
The results in \autoref{tab:main_results} show that closed-source LLMs mostly outperform open-source ones, especially on cross-document event analysis tasks, demonstrating the effectiveness of state-of-the-art proprietary models. 
%cue一下glm
Among them, GPT-5.2-Pro and Claude-4.5-Opus achieve the best performance across nearly all the tasks, while Deepseek-V3.2 and GLM-4.7 rank highest among open-source LLMs.
We also have other interesting findings:
%cross-single分析
%四个task的整体表现(主要focus single-cross),点出一些反常的细节
%下面开始提具体的点

%RAG方法:有效,但不总是有效
\noindent\textbf{RAG can generally improve performance across the four tasks, but the gains are unstable.} 
Specifically, RAG benefits strong LLMs such as Gemini-3-Pro on most of the tasks.
However, for weaker LLMs like Qwen-3-8B, the noisy retrieved documents may degrade the original performance in turn.
How to design specific RAG strategies for event analysis still needs further study.

%2.Reasoning任务:temp,sub上升, cau,core下降
\noindent\textbf{At the MiGUE-Reasoning task, temporal and subevent relations generally benefit from cross-document settings, whereas coreference and causal reasoning exhibit the opposite trend.} 
%This phenomenon is consistently observed in both MiGUE-Reasoning and MiGUE-Induction. 
We conjecture that multi-document contexts may introduce additional temporal signals (e.g., explicit timestamps), which can be easily captured by LLMs.
%are  relatively easy for current LLMs to extract. 
Consequently, temporal-related relations benefit from these additional signals.
%As a result, richer temporal information improves temporal ordering and temporally correlated relations such as subevent detection. 
In contrast, for coreference and causal reasoning, different documents may describe the same event with distinct perspectives. Also, temporally successive events are not necessarily causally related. 
Such heterogeneity may introduces noise and ambiguity, leading to degraded performance in cross-document settings. 
%compared to single-document scenarios.

%1.timespan
\noindent\textbf{At the MiGUE-Induction task,  LLMs are sensitive to start times of events but exhibit limited understanding of their time spans.} 
We observe from the subtask of interval-based temporal ordering that 
%MiGUE-Temporal Order highlights this gap: 
most LLMs perform significantly worse on end-time ordering than on start-time one, with the accuracy often dropping by nearly half. 
This suggests that identifying relative event start times is much simpler for current LLMs, whereas reasoning about event durations remains challenging.

%Causal
\noindent\textbf{Causal tasks remain challenging for current LLMs.} Most of these LLMs achieve unsatisfactory performance 
%As evidenced by MiGUE-
on the tasks of causal reasoning and causal graph induction. We conjecture that one reason is the misalignment between model-internal causal representations and human causal understanding. Moreover, current causal definitions in event analysis are often confined to language-based descriptions and lack rigorous mathematical formalization. We leave the exploration of 
%Future research should explore 
more fine-grained probabilistic causal modeling at the schema level instead of merely textual logical formulations in event analysis as important future work.
%moving beyond purely textual logical formulations.

%%********************时序分析实验图***************
\begin{figure}[t]
  \centering
  \includegraphics[width=\linewidth]{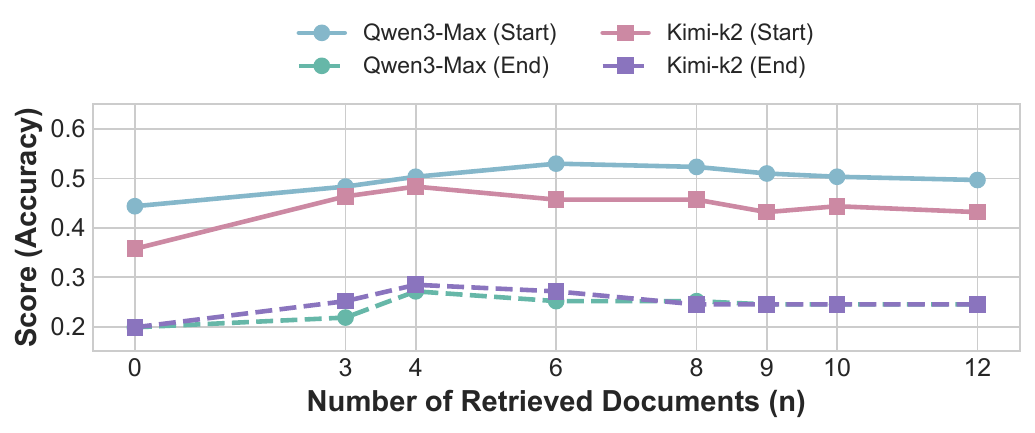}
  %\caption{Model Performance vs. Retrival Document Number}
  \caption{Accuracy on MiGUE-Induction (Temporal Order) with different numbers of retrieved documents.}
    \label{analysis-retrival}
  %\Description{...}
\end{figure}
%%********************时序分析实验 end***************

%%*******************长度分析实验***************
\begin{figure}[t]
  \centering
  \includegraphics[width=\linewidth]{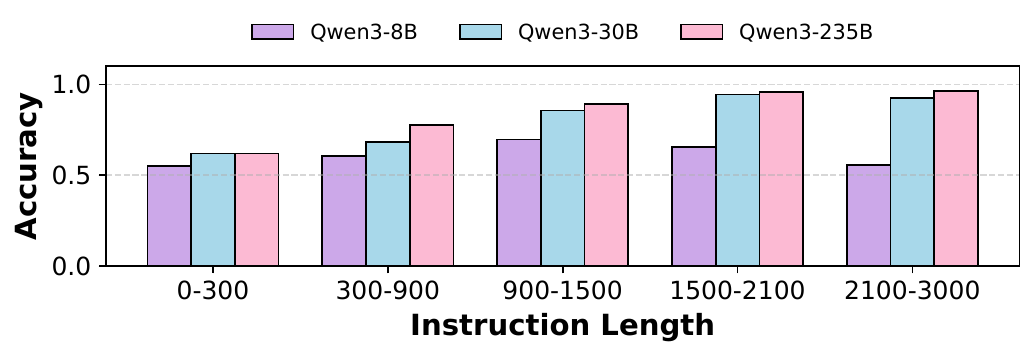}
  %\caption{Model Performance vs. Retrival Document Number}
  %\caption{Impact of instruction length on differen scaling model performance on MiGUE-Reasoning (Subevent).}
  \caption{Accuracy on MiGUE-Reasoning (Subevent) with different lengths of input instructions.}
  %Impact of instruction length on differen scaling model performance on .}
    \label{analysis-scaling}
  %\Description{...}
\end{figure}
%%*******************长度分析实验 end***************
\subsection{Detailed Analysis}

%分析实验
%2.输入长度()------cross 取点 base model ,scale?
%3.错误分析(Induction)
%1.rag- doc num(先跑) Induction-qwenmax 
%\subsubsection{Mistake Analysis on End-Order}
%\noindent\textbf{Temporal-Order vs. Number of Retrieved Documents.}
\noindent\textbf{Analysis on the Number of Retrieved Documents.}
To further analyze this effect of additional relevant documents on temporal reasoning, we evaluate Qwen3-Max and Kimi-K2 under UltraRAG with different numbers of retrieved documents.
Figure \ref{analysis-retrival} shows that both models exhibit a rise–fall–stabilization pattern for start-ordering and end-ordering accuracies as the number of retrieved documents increases. Moderate retrieved documents improve performance, whereas excessive documents degrade accuracy, possibly due to noise accumulation and context dilution. 
End-ordering performance consistently remains lower than start-ordering, revealing the deficiencies of LLMs in modeling event time spans. 
%Qwen3-Max peaks at n = 6, while Kimi-K2 reaches peak performance at n = 4 for both tasks. This divergence suggests model-specific tolerance to retrieval-induced noise, with Kimi-K2 being more sensitive to excessive context and thus better suited for concise RAG inputs.

%\noindent\textbf{Scaling Analysis on Subevent Reasoning}
\noindent\textbf{Analysis on the Length of Input Instructions.}
We further analyze the scaling behavior of LLMs with different lengths of input instructions
%models with different parameter sizes 
on subevent reasoning. As shown in \autoref{analysis-scaling}, Qwen3-235B exhibits steady performance gains as the instruction length increases, 
%peaking at 1500–2100 tokens and remaining stable thereafter, 
indicating strong capabilities of long-context understanding. 
%to long contexts.
For comparison, Qwen3-30B/8B also benefits from longer inputs but reaches its peak in the 1500–2100 / 900–1500 range, followed by a slight decline.
%suggesting diminishing returns beyond a moderate context length.
%For qwen3-8b, performance improves only up to 900–1500 tokens before degrading with further length increases, reflecting greater sensitivity to extended contexts.
This suggest that 
larger models benefit from richer contextual evidence, 
whereas smaller models saturate earlier and degrade under extended inputs.

\section{Conclusion}
In this work, we present a comprehensive benchmark called MiGUE-Bench for multi-granularity event analysis, 
which covers the tasks ranging from fine-grained event detection and reasoning to cross-document event structure induction and forecasting, 
%To support the automatic construction of MiGUE-Bench, we also build an LLM-driven pipeline called MiGUE-Pipeline for scalable event data generation.
together with an LLM-driven automatic pipeline named MiGUE-Pipeline for scalable event data construction. 
%MiGUE-Bench enables a holistic evaluation of LLMs across multiple levels of event-centric reasoning, covering tasks ranging from fine-grained event detection and reasoning to cross-document event structure induction and forecasting. 
Through extensive experiments on state-of-the-art LLMs and RAG methods, we show that despite their strong general capabilities, these models still exhibit notable limitations in multi-granularity event analysis. We hope that MiGUE-Bench and MiGUE-Pipeline can serve as standardized testbeds for diagnosing the capability boundaries of LLMs in event analysis and inspire future advances in this research field.

%%
%附录部分
%\appendix
%No-Occurrence, Negated-Claims, Assumption, Abstraction, Named-Entity,} and {\itshape Narrative
%\section{Description of Event Annotation Error Types}
%想了一下还是放正文算了...

%%
%致谢
%

\begin{acks}
This work was supported by Noncommunicable Chronic Diseases-National Science and Technology Major Project (No. 2023ZD0501806), Sichuan Science and Technology Program (No. 2025ZNSFSC1488)
, Fundamental Research Funds for the Central Universities (No. ZYGX2025XJ041), and CIPS-SMP-Zhipu Large Model Fund (No. CIPS-SMP20250314).

%\grantsponsor{NCDNSTMP}{Noncommunicable Chronic Diseases-National Science and Technology Major Project}{}
%\grantnum{NCDNSTMP}{2023ZD0501806}

%\grantsponsor{SCSTP}{Sichuan Science and Technology Program}{}
%\grantnum{SCSTP}{2025ZNSFSC1488}

%\grantsponsor{FRFCU}{Fundamental Research Funds for the Central Universities}{}
%\grantnum{FRFCU}{ZYGX\allowbreak2025XJ041}

%\grantsponsor{CIPSSMPZHIPU}{CIPS-SMP-Zhipu Large Model Fund}{}
%\grantnum{CIPSSMPZHIPU}{CIPS-SMP20250314}
\end{acks}

%%
%参考文献部分
\bibliographystyle{ACM-Reference-Format}
\bibliography{main_ref}

\end{document}